\documentclass[11pt, onecolumn, doublespacing, lineo, a4paper, peerreview]{IEEEtran}
\usepackage{lineno}
\usepackage{geometry} 
\usepackage{tikz}
\usepackage{xcolor}
\usepackage{color}
\usepackage{graphicx}
\usepackage[cmex10]{amsmath}
\usepackage{algorithm2e}
\usepackage{breqn}
\usepackage{subfigure}
\usepackage{url}
\usepackage{booktabs}

\newcommand{\ie}{i.e.}

\newcommand{\eg}{e.g.}

\renewcommand{\d}[1]{{\mbox{\boldmath$#1$}}}
\newcommand{\m}[1]{{\mbox{{\fontencoding{T1}\sffamily\slshape{#1\/}}}}}

\newcommand{\ical}[1]{{\mbox{\usefont{OT1}{pzc}{m}{it}{#1}}}}
\DeclareMathOperator*{\argmin}{argmin}

\begin{document}
\title{Archetypal Analysis for Sparse Representation-based Hyperspectral Sub-pixel Quantification}

\author{\IEEEauthorblockN{Lukas Drees, Ribana Roscher, Susanne Wenzel}\\
\IEEEauthorblockA{Institute of Geodesy and Geoinformation, 
University of Bonn, 
Germany, 53115 Bonn\\
s7ludree@uni-bonn.de, ribana.roscher@uni-bonn.de, wenzel@igg.uni-bonn.de}}

\maketitle

\begin{abstract}
The estimation of land cover fractions from remote sensing images is a frequently used indicator of the environmental quality.
This paper focuses on the quantification of land cover fractions in an urban area of Berlin, Germany, using simulated hyperspectral EnMAP data with a spatial resolution of 30m$\times$30m. 
We use constrained sparse representation, where each pixel with unknown surface characteristics is expressed by a weighted linear combination of elementary spectra with known land cover class. 
We automatically determine the elementary spectra from image reference data using archetypal analysis by simplex volume maximization, and combine it with reversible jump Markov chain Monte Carlo method.
In our experiments, the estimation of the automatically derived elementary spectra is compared to the estimation obtained by a manually designed spectral library by means of reconstruction error, mean absolute error of the fraction estimates, sum of fractions, $R^2$, and the number of used elementary spectra.
The experiments show that a collection of archetypes can be an adequate and efficient alternative to the manually designed spectral library with respect to the mentioned criteria.
\end{abstract}

\IEEEpeerreviewmaketitle

\section{Introduction}\label{introduction}
The estimation of the degree of imperviousness as an indicator of environmental quality is subject of current research towards a time and cost efficient monitoring of urban areas \cite{Weng2012}.
Due to increasing land consumption in cities in the recent years, which has negative effects on the natural water cycle, the monitoring of land use in those areas is important \cite{wessolek2001bodenuberformung}.

Remote sensing data, such as imaging spectroscopy, builds a valuable basis to comprehensively map urban areas and quantify the imperviousness based on the spectral information (\eg, \cite{Huang2014,Roessner2011}).
Especially, hyperspectral imagery is a suitable source for mapping of such areas, because it offers a high spectral separability of different materials.
However, generally, the temporal and spatial resolution is limited in comparison to sensors with lower spectral resolution.
These limitations are partially overcome with the launch of missions such as Environmental Mapping and Analysis Program (EnMAP), which increases the availability of hyperspectral data and the temporal resolution \cite{Guanter2015}.
Nevertheless, due to its spatial resolution, the provided data is mainly characterized by spectrally mixed pixels, which demands sophisticated sub-pixel quantification approaches in order to estimate the fraction of various land cover classes in each pixel.

In this context, several approaches have been developed comprising regression approaches \cite{Okujeni2016,Priem2016}, probabilistic classification methods \cite{Rosentreter2017,Suess2014,zhu2012kernel}, and the usage of spectral libraries for spectral mixture analysis \cite{Somers2011, Powell2007}.
An overview of a wide variety of unmixing approaches can, for example, be found in \cite{Bioucas-Dias2012}.
While the latter approach needs a spectral library containing elementary spectra of known materials, the first two approaches also require mixed spectra for learning an appropriate model. 
These mixed spectra can be derived from the image using information about known mixed pixels, or from synthetically mixed pixel, as it has been presented in \cite{Rosentreter2017} and \cite{Okujeni2016}.

When using spectral libraries, a critical step is the extraction of the elementary spectra.
A manual extraction is time-consuming and requires human expert-knowledge and therefore, automatic extraction techniques have been an active field of research during the past decade (\eg, \cite{Bioucas-Dias2012,Veganzones2008}).
Most of the algorithms rely on the assumption that the elementary spectra lie on a convex hull or a convex polytope enclosing the data distribution (\eg, \cite{Chan2011,Craig1994}).
Based on this assumption all data samples can be reconstructed by a non-negative linear combination of the elementary spectra.
A promising approach from this group is the so-called archetypal analysis, which searches for extreme points (also known as archetypes) in the data distribution (\eg, \cite{Zhao2017,Zhao2016,Morup2012, Cutler1994}).
Archetypal analysis has already been successfully applied in the field of sport analytics \cite{seth2016probabilistic}, plant phenotyping \cite{romer2012early} or text analysis \cite{seiler2013archetypal}. 
A valuable extension to archetypal analysis is presented by \cite{Heylen2015}, in which extreme points are extracted in the kernel space, enabling an efficient nonlinear unmixing.
Besides the actual determination of elementary spectra, other challenges exist which need to be tackled: The number of elementary spectra is unknown beforehand and thus, a suitable amount of spectra needs to be extracted to make the set representative enough, but also small enough to keep the sub-pixel quantification robust and efficient. 
Moreover, many extraction techniques depend on the initialization and thus, a strategy needs to be defined to ensure a stable result (\eg, \cite{Wang2016,Zortea2009}).

\begin{figure*} [htb]
    \centering
	\subfigure[Part of HyMap image data visualized as RGB-image with the wavelengths R = $640$nm, G = $540$nm and B = $450$nm]{
	\includegraphics[height=0.77\textwidth]{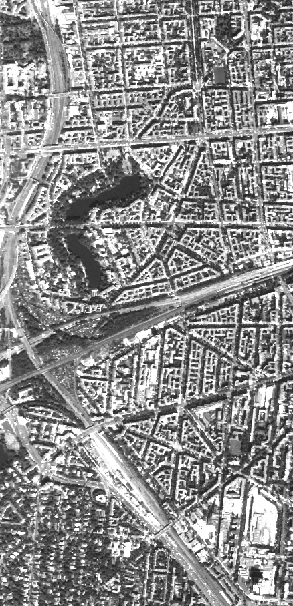}}
	\subfigure[Reference image with four classes \texttt{impervious surface}, \texttt{vegetation}, \texttt{soil} and \texttt{water}]{
	\includegraphics[height=0.77\textwidth]{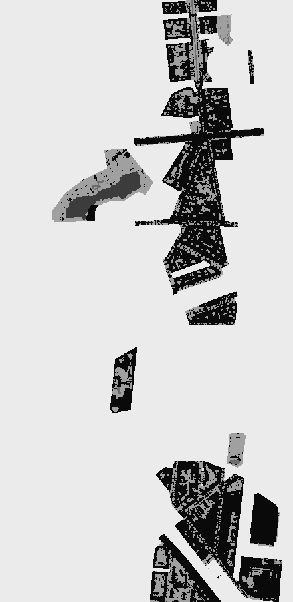}}
	\subfigure[Simulated EnMAP data visualized with the wavelengths as specified above (bands: RGB = 11,5,1)]{
	\includegraphics[height=0.77\textwidth]{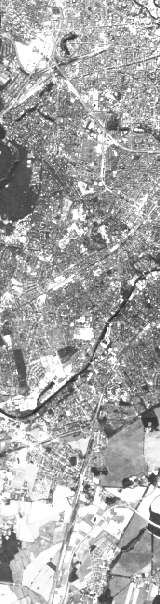}}
    \caption[HyMap - Reference - EnMAP]{Berlin-Urban-Gradient dataset} 
\label{fig:imag}
\end{figure*}

In this paper, we address the challenge of automatically finding a representative set of elementary spectra, including the automatic determination of the number of elements, for sub-pixel quantification.
We perform the sub-pixel quantification using a freely available simulated EnMAP scene of an urban area in Berlin, Germany \cite{Okujeni2016a}\footnote{\url{http://pmd.gfz-potsdam.de/enmap/showshort.php?id=escidoc:1480925}}, aiming at the estimation of a fraction map containing the classes \texttt{impervious surface}, \texttt{vegetation}, \texttt{soil} and \texttt{water}.
In order to determine the class fractions, we use sparse representation with non-negativity, L$_0$-sparsity and sum-to-one constraint.
We exploit archetypal analysis to extract elementary spectra in a fully automatic and unsupervised way.
Moreover, we perform archetypal analysis by simplex volume maximization (SiVM), which states an efficient selection method \cite{Thurau2010,heylen2011non}.

The main contribution of this work is the combination of archetypal analysis with reversible jump Markov chain Monte Carlo method (rjMCMC, \cite{Green1995}) to obtain a spectral library of high representational power, yet a small number of elementary spectra. 
Using this approach, we are able to determine the best set of spectral library elements regarding a pre-defined criteria, as well as the number of elements.
Moreover, the approach of \cite{Heylen2015} is applied to select the archetypes in kernel space, resulting in archetypes lying on the concave hull of the data distribution in the original feature space.
Our experiments confirm that these archetypes are more suitable for sub-pixel quantification than archetypes extracted from the convex hull exclusively.
To illustrate the usefulness of our proposed approach, we analyze various kinds of automatically derived spectral library and compare them to a manually designed spectral library.
Our presented approach is flexible regarding the chosen estimation technique, such that the constrained sparse representation can be replaced by other approaches commonly used in the unmixing community (\cite{Bioucas-Dias2012}), or other constraints such as sparsity induced by L$_1$-norm (\cite{Zhang2015}).
As such, archetypal analysis can be replaced by, e.g., endmember extraction methods presented in \cite{Bioucas-Dias2012}, and deliver the input for the rjMCMC method.

\section{Data}
Our studies are performed using the Berlin-Urban-Gradient dataset \cite{Okujeni2016a}, illustrated in Fig. \ref{fig:imag}. 
The dataset consists of two hyperspectral images of different spatial resolution, two simulated EnMAP scenes of different spectral resolution, a manually designed spectral library, reference land cover information and reference fractions for evaluation, which are explained in more detail in the following paragraphs.

\begin{figure} [htb]
    \centering
	\makebox[0pt]{
        \includegraphics[width=0.6\textwidth]{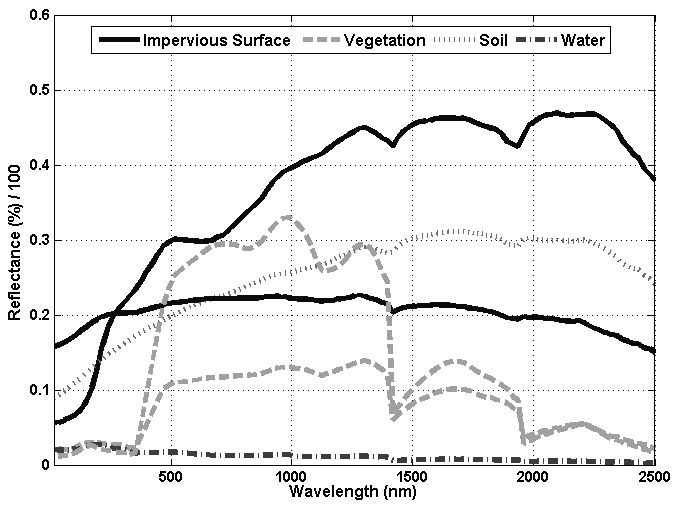}
        }
    \caption[Spectral Library spectra]{Significant spectra of the spectral library (\texttt{Lib}). Asphalt and red clay tile roof in the class \texttt{imp. surface}, grass and tree for \texttt{vegetation}, bare \texttt{soil} and natural \texttt{water}.} 
\label{fig:Libspec}
\end{figure}

\begin{table}[thb]
\renewcommand{\arraystretch}{1.3}
\centering
\caption[Composition of spectral library]{Composition of 75 spectra in the manually designed spectral library (\texttt{LibCom})} 
\begin{tabular}[b]{c c c c}
\toprule
\texttt{Imp. surface} & \texttt{Vegetation} & \texttt{Soil} & \texttt{Water}\\
\midrule
39 & 31 & 4 & 1\\
\bottomrule
\end{tabular}\\[1em]
\label{tab:Lib}
\end{table}

\subsubsection{HyMap Data}
\label{sec:hypmap}
The dataset contains two hyperspectral images, one with a spatial resolution of $3.6$m and one with a spatial resolution of $9$m, whereby we use the higher resolution image in our work.
Both images were acquired with the Hyperspectral Mapper (HyMap), and serve as basis for the extraction of the archetypes.
It offers a high variety of urban land use and land cover patterns in the study site Southwest of Berlin, Germany.
The external conditions were a cloudless sky at solar noon and a minimal possible altitude (maximal altitude according to the second lower resolution image).
The HyMap image consists of 126 spectral bands, however, 15 noisy bands were removed resulting in 111 bands used for this work. 
Moreover, other preprocessing steps were performed, as encompassed system correction \cite{cocks1998hymaptm}, atmospheric correction and parametric geocoding \cite{richter2002geo}.
The observed wavelengths range from 0.45$\mu$m to 2.5$\mu$m, showing a high spectral information diversity, which enables a detailed analysis of the urban structure.
In addition to the $3.6$m-resolution HyMap image, a manually derived reference image containing $112.690$ reference pixel is provided with valuable spectra, each labeled with one of the four land cover classes: \texttt{impervious surface}, \texttt{vegetation}, \texttt{soil} and \texttt{water}. 
The reference information is manually obtained by using digital orthophotos and cadastral data.
For our experiments, the set of elementary spectra is extracted from this data to obtain the class membership of the elementary spectra, however, the extraction can be performed on unlabeled data with limited human user interaction.

\subsubsection{Simulated EnMAP Data}
\label{sec:simEn}
EnMAP is German hyperspectral satellite mission, which will start not earlier than 2018, with a focus on Earth environmental observations in a global scale.
Based on the HyMap data, an EnMAP scene was simulated using EnMAP end-to-end simulation tool (EeteS, \cite{Segl2012}) with two different spectral resolutions (111 and 244 bands). 
Just as the HyMap data, both EnMAP images have a spectral resolution ranging from $0.45$$\mu$m to $2.5$$\mu$m, where we use $111$ bands for a better comparability.
The spatial resolution of $30$m is lower than the resolution of the HyMap scene, and thus the mixing of land cover classes are more apparent. 
For evaluation purposes, $1495$ EnMAP pixels were obtained from the simulation tool, containing the fractions of the land cover classes ranging from $0$ to $100\%$.

\subsubsection{Manually Designed Spectral Library}
The spectral library is a manually designed collection of $75$ pure spectra obtained from the HyMap image. 
The spectra contain different land cover classes with $39$ impervious surface spectra (different types of roof, pavement, tartan, pool water), $31$ vegetation spectra (grass, tree), $4$ soil spectra (uncovered ground, sand) and one natural water spectrum. 
All together, it is a balanced library for urban structures with $39$ impervious and $36$ pervious spectra. 
All spectra in the library can be assigned to a hierarchical urban classification scheme, which was developed by \cite{heiden2007}. 
First, an initial collection of $300$ spectra was selected by expert knowledge.
Afterwards a two-step filtering was performed, in which the variability between the spectra is maximized in consideration of spectral variability of materials, brightness and shading effects. 
Moreover, in an iterative process those spectra for the final subset were selected that best describes the spectral diversity in a specific neighborhood. 
More details can be found in \cite{Okujeni2016a}.

\section{Methods}
The following section describes the sub-pixel quantification using sparse representation and archetypal analysis.
Archetypal analysis determines the extreme points of the data distribution, which are used within the sparse representation approach to estimate the fractions of land cover classes in each pixel.
We have given a $(M\times N)$-dimensional data matrix $\m{X}$, in which $N$ is the number of $M$-dimensional reference pixels $\m{X} = [\d{x}_1, \ldots,\d{x}_N]$ with given land cover class $c_n$.
Moreover, we have given a test set of $1495$ data samples ${}^{\text{T}}\m X = [\d{x}_1, \ldots,\d{x}_T]$ with known land cover fractions ${}^{\text{T}}\d f_t$ for evaluation purposes, as presented in Sec. \ref{sec:simEn}.

\subsection{Sparse Representation}
In order to determine the sub-pixel fractions, we use sparse representation with non-negative least squares optimization.
In terms of sparse coding, a sample $\d x$ is represented by a weighted linear combination of a few elements taken from a $\left(M \times D\right)$-dimensional dictionary $\m D$, such that \mbox{$\d x = \m D \d \alpha + \d \gamma$} with $\Vert \d \gamma \Vert_2$ being the reconstruction error.
The dictionary $\m D = [\d x_1, \ldots, \d x_d, \ldots, \d x_D]$ contains elementary spectra, such that this approach is identical to the linear mixing model.
The coefficient vector, comprising the activations, is given by $\d \alpha$.
The activations are interpreted as class fractions for sub-pixel quantification.
The optimization problem for the determination of optimal $\d{\hat\alpha}$ is given by
\begin{equation}
\label{eq:SR}
	\d{\hat\alpha} = \argmin~ \Vert \m D\d \alpha - \d x \Vert_2,
\end{equation}
\begin{equation}
\label{eq:SR2}
	\quad \text{subject to} \quad \d{\alpha} \geq \d 0, ~\sum_d \d{\alpha}_d = 1, ~\Vert \d \alpha \Vert_0 \leq W
\end{equation}
where the terms in \eqref{eq:SR2} are the non-negativity constraint, the sum-to-one constraint, and the sparsity constraint. 
Generally, non-negativity alone leads to a sparse solution, however, in order to ensure a strict fulfillment of the sparsity constraint we use a backward selection procedure in which the activations with the smallest values are set to zero in a greedy manner.

\subsection{Archetypal Analysis}
Archetypal analysis is a suitable method to determine the elements of $\m D$, where each archetype serves as one dictionary element. 
The extraction of the archetypes, collected in a $(M \times K)$-dimensional matrix of $ \m {A} = [\d {a}_1, \ldots, \d {a}_k, \ldots, \d {a}_K] $, $k=1,\ldots,K$, is carried out by SiVM, which is an efficient method to determine the archetypes of the data distribution. 
This approach is aiming on an approximation of the convex hull, where all archetypes are located on it. 

In order to find the first archetype, the approach is initialized with a random or pre-defined vector $\d a_0$.
The pixel with maximal distance to $\d a_0$, is defined as first archetype $\d{a}_1$, defined by
\begin{equation}
	\d{a}_1 = \operatorname*{arg\,max}_n \ical{d(}\d a_0,\d x_n\ical{)},  \label{eq:sivm3}
\end{equation}
with $\ical{d(}\cdot, \cdot\ical{)}$ being the Euclidean distance function between the spectral features of the archetype $\d a_0$ and the pixel $\d x_n$.
Further archetypes are specified sequentially, such that the volume of the simplex becomes maximized with each additional archetype. 
Since the volume operation is too computational intense, instead the archetypes are selected to have maximum distance to all previously detected ones, using
\begin{equation}
	\d{a}_M = \operatorname*{arg\,max}_{n} \sum_k \ical{d(}\d a_k, \d x_n\ical{)}.  \label{eq:sivm5}
\end{equation}
The stopping criterion is generally chosen to be the number of archetypes.

We further use the approach of \cite{Heylen2015} and transform $\m X$ into kernel space using a Gaussian radial basis function kernel with a hyperparameter $\sigma$ describing the width of the Gaussian kernel.
In this way, archetypes are selected which lie on the the concave hull rather than the convex hull.

The disadvantage of archetypal analysis is that the final set depends on the initialization point, and as a result, there is no unique solution to the final set. 
Especially, if the number of archetypes in the dictionary is low, various solutions lead to significantly different accuracies.
Moreover, the number of archetypes is generally not know beforehand, and depends on the number of informative dimensions and the variability of the data. 

\subsection{Reversible Jump Markov Chain Monte Carlo Method}
To overcome this problem, we propose to use an optimization procedure to find the best set of archetypes from a large set of pre-selected ones, called the initial set.
Under the assumption that a suitable archetypal set is able to represent the test set ${}^{\text{T}}\m X$ with a low reconstruction error, our task is to find the set of archetypes $\m D = \m A$ which minimizes the energy
\begin{equation}
	\ical U(\m D) = \Vert \d \gamma \Vert_2\,\ ,
\end{equation}
where $\d \gamma$ is obtained by using the current set of archetypes as dictionary $\m D$.
The energy $\ical U$ is a complex function with rough landscape and unknown dimensionality due to the unknown number of archetypes.
Therefore, we optimize with rjMCMC coupled with simulated annealing to find the global optimum.
Introducing the temperature parameter $R$, the optimizer is given by
\begin{equation}
\displaystyle
	\widehat{\mathcal{\m D}}
	= \displaystyle\argmin_{\mathcal{\m D}}\  \frac{\ical U\left(\mathcal{\m D}\right)}{R_k},
	\ \lim_{k\rightarrow \infty} R_k = 0\ .
\vspace{-0.1cm}
\end{equation}
While MCMC is dedicated to sample from complex functions, simulated annealing allows to make a point estimate of its global optimum.
Using simulated annealing we create a Markov chain, such that the samples explore the whole state space in the beginning and gradually concentrate around the global optimum of the energy function $\ical U$.
In this way we avoid trapping into local optima, as it is usually the case for greedy algorithms.
We use the so-called birth an death algorithm \cite{geyer*94:simulation} to sample from a restricted sample space of possible sets of archetypes, which turns out to be a special type of Green's rjMCMC sampler \cite{Green1995}.
The sample space is restricted by a Poisson prior on the expected number of archetypes, as presented in \cite{Bulatov2017}.
We further introduce an upper bound on the the number of selected archetypes.

\section{Experiments}
\subsection{Experimental Setup}
\label{sec:setup}
In our experiments, the simulated EnMAP data is reconstructed by sparse representation with the before mentioned constraints using the following libraries: 
\begin{enumerate}
	\item the manually designed spectral library (\texttt{LibCom}),
	\item an optimized set of the manually designed spectral library \texttt{LibRed} (using rjMCMC),
	\item a library containing $40$ extracted archetypes in the original feature space initialized by the mean vector (\texttt{AA-M-Lin}),  
	\item a library containing $75$ extracted archetypes in the kernel space initialized by the mean vector (\texttt{AA-M}), 
	\item a library containing $75$ extracted archetypes in the kernel space initialized by a random vector (\texttt{AA-Rand}),
	\item a library containing an accumulated set of multiple single sets of archetypes, obtained in the kernel space with random initializations (\texttt{AA-Full}), and
	\item an optimized set of \texttt{AA-Full} denoted by \texttt{AA-Opt} (using rjMCMC).
\end{enumerate}
We compare our results to the regression and classification methods presented in \cite{Rosentreter2017}, namely support vector machine (\texttt{SVM}), import vector machine (\texttt{IVM}), support vector regression (\texttt{SVR}), and multi-output support vector regression (\texttt{MSVR}).
The aim of the experiments is to show the suitability of the spectral libraries for sub-pixel quantification.
Moreover, the goal is to show that the set of automatically derived spectral libraries using archetypal analysis and rjMCMC achieve similar results than the manually designed spectral library.
The sub-pixel quantification is evaluated with the given reference mixing fractions.
Additionally, the number of used archetypes is evaluated and discussed.
We run all experiments including a random component 10 times and report the average result and standard deviation.

\begin{figure} [htb]
    \centering
        \includegraphics[width=0.7\columnwidth]{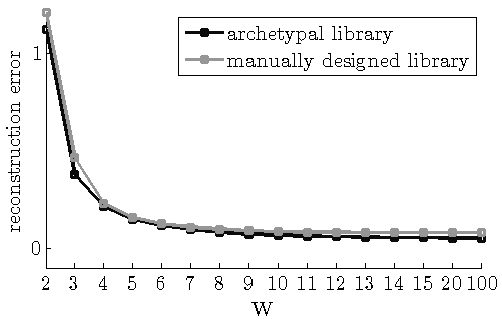}
    \caption[]{Reconstruction error evaluated on the test set for different sparsity values $W$.}
\label{fig:recElbow}
\end{figure}

As pre-processing step, the dataset is outlier-cleaned using the local outlier factor approach \cite{Breunig2000}, using $10$ neighbors and a quantile of $0.95$ as threshold on the pairwise Euclidean distances.  
The width of the Gaussian kernel $\sigma$ is chosen to be $0.5$ of the average standard deviation over all bands.
In order to choose a suitable sparsity value $W$, we use Elbow method to analyze the reconstruction error of the test set obtained by the manually designed library as well as the automatically derived library containing $75$ archetypes. 
Fig. \ref{fig:recElbow} shows that the reconstruction error will not significantly decrease for $W>7$, and thus we choose this value as upper bound for the sparsity constraint.
For rjMCMC we choose the prior on the expected number of archetypes to be $75$ when using all labeled data, and $40$ when using the manually designed spectral library. 
The upper limit of archetypes is chosen to be $150$.

\subsection{Evaluation}
We use several statistics for evaluation.
First, the reconstruction error of the sparse representation is provided, indicating the ability of the elementary spectra to represent the EnMAP pixels.
Moreover, the result of the sparse representation is evaluated by a comparison between reference and actual estimated fraction coefficients. 
This is done by means of the overall and class-wise mean absolute error ($\operatorname{MAE}$) between reference and actual coefficients.
\begin{equation}
\operatorname{MAE} = \dfrac{1}{T} \sum_t^T\left| ({}^{\text{T}}\d{f}_{t} - {}^{\text{I}}\d{f}_{t}) \right|,
\end{equation}
with ${}^{\text{T}}\d{f}_t$ being the reference fractions, ${}^{\text{I}}\d{f}_t$ being the estimated fractions, and $T$ being the number of evaluated pixels. 
We obtain ${}^{\text{I}}\d{f}_t$ by summing up all coefficients belonging to elementary spectra of the same class. 
The smaller the $\operatorname{MAE}$, the better the reconstruction.
Moreover, the root mean square error ($\operatorname{RMSE}$) is provided, which is defined by
\begin{equation}
\operatorname{RMSE} = \dfrac{1}{T} \sum_t^T \sqrt{({}^{\text{T}}\d{f}_{t} - {}^{\text{I}}\d{f}_{t})^2}.
\end{equation}
Besides this, we provide the coefficient of determination ($R^2$) for each class. 
It is calculated as the squared correlation coefficient between ${}^{\text{T}}\d{f}_t$ and ${}^{\text{I}}\d{f}_t$.

\subsection{Results and Discussion}

\begin{table*}[thb]
\scriptsize
\renewcommand{\arraystretch}{1.3}
\centering
\caption{Evaluation results of EnMAP sub-pixel quantification. Overall (\O) and class-wise $\operatorname{MAE}$ (in brackets computed without zero-reference fractions) and overall $\operatorname{RMSE}$. Sparse representation-based estimations with various obtained spectral libraries (explanations see Sec. \ref{sec:setup}) are compared to regression (SVR, MSVR) and classification methods (IVM, SVM). The standard deviation values are indicated by $\pm$.} 
\begin{tabular}[b]{c c c cccc c c}
\toprule
Data & Approach & Amount & \multicolumn{4}{c}{Class-wise MAE [\%]} & \O~MAE [\%] & \O~RMSE [\%] \\
\cmidrule{4-7}
& & of elements & \texttt{Imp. surface} & \texttt{Vegetation} & \texttt{Soil} & \texttt{Water} & & \\
\midrule
$\m X_{\text{lib}}$ & \texttt{SVM} & 75 & 14.81 (15.22) & 16.97 (15.97) & 04.84 (19.01) &  04.25 (66.33) &  10.22 & 17.36 \\
$\m X_{\text{lib}}$ & \texttt{IVM} & 75 & 15.21 (15.91) & 15.31 (15.59) & 02.09 (20.98) &  01.98 (35.51) &  08.65 & 15.63  \\
$\m X_{\text{lib}}$ & \texttt{SVR} & 75 & 11.73 (11.51) & 12.20 (11.92) & 08.36 (12.78) &  07.95 (07.68) &  10.06 & 13.74  \\
$\m X_{\text{lib}}$ & \texttt{MSVR} & 75 & 11.33 (11.45) & 10.99 (11.49) & 02.17 (13.45) &  03.19 (09.70) &  06.92 & 11.32 \\
\midrule
$\m X_{\text{lib}}$ & \texttt{LibCom} & 75 & 13.00 (13.79) & 09.61 (09.87) & 02.01 (14.37) & 07.96(07.88) & 08.15 & 12.66 \\
$\m X_{\text{lib}}$ & \texttt{LibRed} & 44.4 & 14.00 (14.86) & 10.58 (10.85) & 02.23 (15.07) & 07.46 (09.07) & 08.57 & 13.68 \\
 &  & $\pm$ 9.5 & $\pm$ 1.45 (1.57) & $\pm$ 0.35 (0.39) & $\pm$ 0.64 (0.61) & $\pm$ 2.00 (1.71) & $\pm$ 0.96 & $\pm$ 1.22\\
\midrule
$\m X$ & \texttt{AA-M-Lin} & 40 & 25.09 (26.23) & 15.05 (15.70) & 01.62 (20.38) & 33.00 (09.65) & 18.69 & 23.65 \\
$\m X$ & \texttt{AA-M} & 75 &  25.50 (27.00) &  22.44 (22.93) &  01.80 (02.04) &  08.87 (20.39) & 14.65 & 20.01 \\
$\m X$ & \texttt{AA-Rand} & 75 & 21.05(21.58) & 17.86 (18.27) &  01.77 (15.69) &  08.52 (26.29) & 12.30 & 18.17 \\
 &  & $\pm$ 0  & $\pm 3.11 (4.13)$  & $\pm 3.05 (3.10)$  & $\pm 0.06 (3.11)$  & $\pm 4.07 (27.91) $ & $\pm 2.00$  & $\pm 1.22$ \\
$\m X$ & \texttt{AA-Full} & 142 & 49.06 (52.71) & 32.50 (33.20) & 02.38 (26.16) & 2.84 (68.13) & 21.70 & 31.45 \\
$\m X$ & \texttt{AA-Opt} & 81.7 & 16.29 (16.21) & 14.21 (14.70) & 03.14 (24.09) & 07.00 (09.73) & 10.16 & 16.08 \\
 &  & $\pm$ 13.6 & $\pm$ 0.69 (0.76) & $\pm$ 0.65 (0.60) & $\pm$ 0.60 (2.91) & $\pm$ 1.50 (1.38) & $\pm$ 0.51 & $\pm$ 0.47\\
\bottomrule
\end{tabular}\\[1em]
\label{tab:ergeb}
\end{table*}

\begin{figure*} [htb]
    \centering
    	\includegraphics[width=1\textwidth]{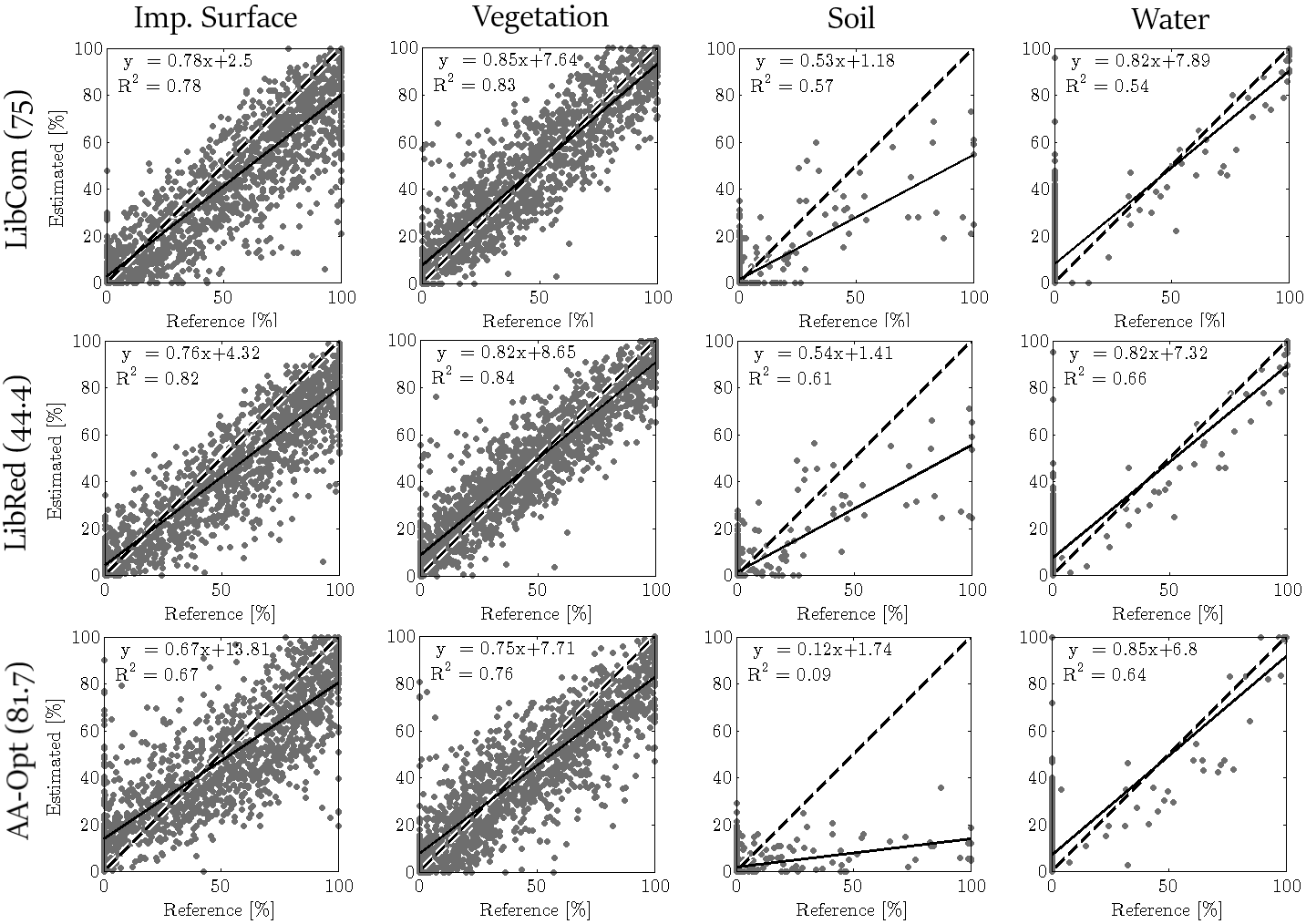}
    \caption[Scatterplot results]{Scatter plots obtained from \texttt{LibCom} (top row), \texttt{LibRed} (middle row)  and \texttt{AA-Opt} (bottom row) representing the class-wise fraction estimates of \texttt{impervious surface}, \texttt{vegetation}, \texttt{soil} and \texttt{water}, opposed to the reference fractions. A correct estimated pixel lies on the dashed line. The solid line represents the respective least-square regression line for the scatter points. The formula for the regression line the and coefficient of determination ($R^2$) is also given in each plot.} 
\label{fig:scat}
\end{figure*}

\begin{figure*} [htb]
    \centering
    	\subfigure[Full manually designed spectral library \texttt{LibCom}]{\includegraphics[width=0.47\textwidth]{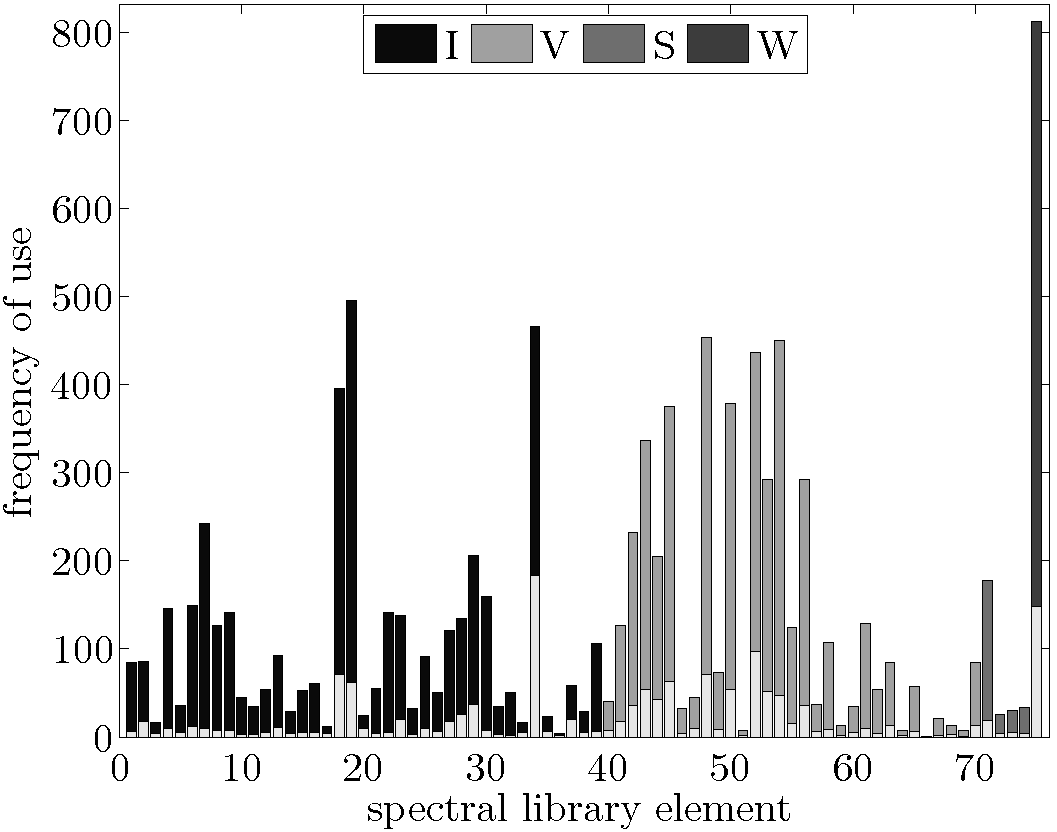}}
         \subfigure[Reduced manually designed spectral library \texttt{LibRed}]{\includegraphics[width=0.47\textwidth]{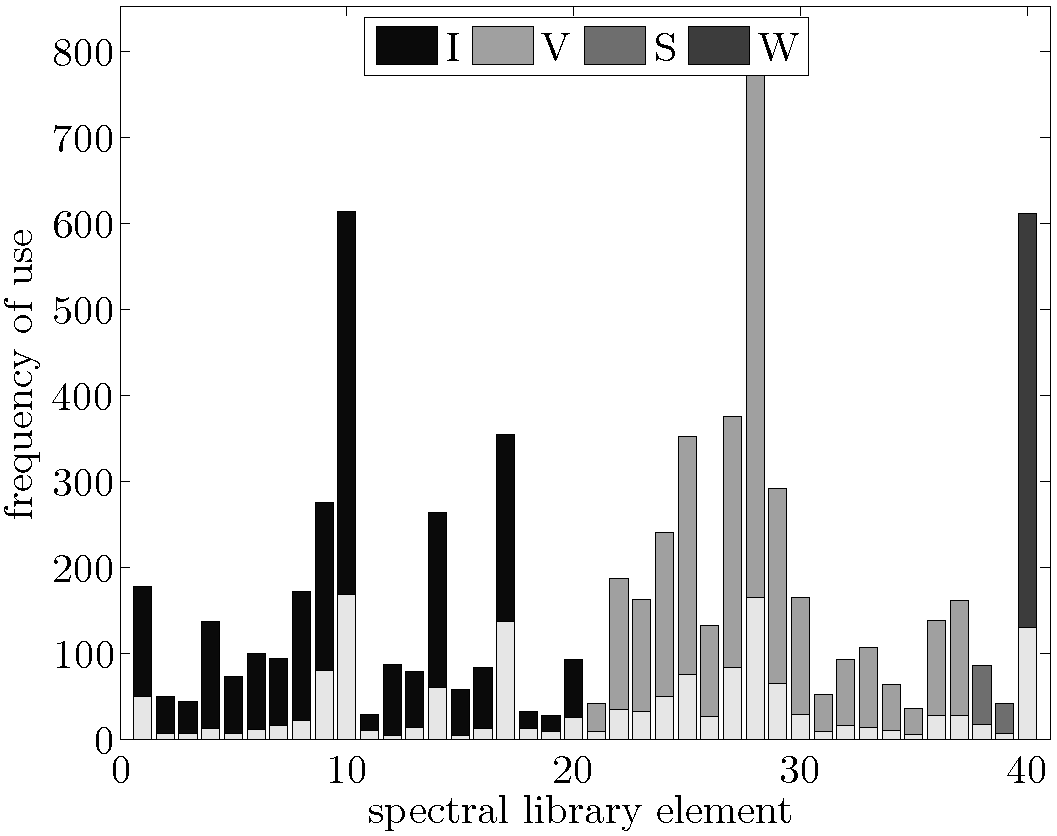}}
    \caption[How often a basis elements is used for EnMAP reconstruction]{The frequency of the usage of elementary spectra for the reconstruction of the reference EnMAP pixels. Different colors indicate the class membership of the reference pixels (I: \texttt{Imp. surface}, V: \texttt{Vegetation}, S: \texttt{Soil}, W: \texttt{Water}). The maximal possible height of a bar is $1493$, if the elementary spectrum is involved in the reconstruction of all reference EnMAP pixels. The lower bright part of each bar specify the sum of the fractions (\ie, the estimated coefficients) over all pixel.} 
\label{fig:reconstruct}
\end{figure*}

Table \ref{tab:ergeb} presents the results of the EnMAP sub-pixel quantification. 
In the first block, the results of \cite{Rosentreter2017} are shown for comparability with regression and classification results.
The middle block shows the results obtained by the full and reduced manually designed spectral library.
The bottom part of the table presents the results obtained by archetypal analysis.

Overall, the $\operatorname{MAE}$ values show that all spectral libraries which are obtained from the full manually designed library $\m X_{\text{lib}}$ achieve similar and satisfactory results. 
It can be seen that in comparison to the results of the regression and classification approaches presented in \cite{Rosentreter2017}, sparse representation with the mentioned constraints show equivalent results.
Both spectral libraries have a small average $\operatorname{MAE}<9\%$, which is also obtained by \texttt{IVM} and $\texttt{MSVR}$.
Remarkably, \texttt{LibCom} and \texttt{LibRed} show similar results, however, with \texttt{LibRed} having much fewer elements.
\texttt{LibCom} has slightly lower class-wise $\operatorname{MAE}$ for the classes \texttt{impervious surface}, \texttt{vegetation} and \texttt{soil}, where \texttt{LibRed} obtains better results for \texttt{water}. 
This indicates that the presented rjMCMC method is able to find a suitable subset of elementary spectra having nearly the same representational power than the manually designed library with all elements.
Moreover, it can be stated from this result that the full spectral library contains redundant elementary spectra which can be discarded for sub-pixel quantification.

These findings are underlined by Fig. \ref{fig:scat}, which shows the scatter plots representing the $1495$ class-wise fraction estimates opposed to the reference fractions.
It can be observed that \texttt{soil} and \texttt{water} have a high proportion of $0\%$ reference fractions and lesser \textgreater$0\%$ fractions than \texttt{impervious surface} and \texttt{vegetation}. 
The \texttt{impervious surface} scatter plot show that for \texttt{LibCom} the two lines intersect nearly at the 15\% point on the x-axis. 
This means that the estimated fractions are usually too small for all pixels which have a degree of impervious surfaces over $15\%$, and otherwise too high for smaller values than the intersection point. 
In comparison to \texttt{LibRed}, the reference values are also mostly underestimated, but slightly better than \texttt{LibCom}.
The \texttt{soil} scatter plots have large discrepancies between the true 1:1 line (dashed line) and the estimated least-squares regression line (solid line). 
Besides this, there is a high amount of $0\%$ fractions, which are estimated with up to $40\%$ when using the \texttt{LibCom} spectra and $30\%$ fractions when using \texttt{LibRed} spectra. 
Furthermore, all pixels with \texttt{soil}-fractions are clearly underestimated equally. 
This is also the case for the regression and classification approaches presented in \cite{Rosentreter2017}.
We assume that the small number of elementary spectra is insufficient for the spectral diversity of \texttt{soil} pixels. 
Finally, the described observation with high estimated $0\%$ fractions also occurs in the \texttt{Water} class with values over $50\%$. 

Figure \ref{fig:reconstruct} illustrates the frequency of the usage of each spectrum for reconstruction in order to determine the fractions of $1493$ EnMAP pixels.
The results obtained by \texttt{LibCom} are shown on the left and the results for \texttt{LibRed} on the right side.
Moreover, the lower bright part of each bar indicates the sum over all fractions from which the total proportion of each land cover class in the study area can be derived.
\texttt{LibCom} shows that there are significant elementary spectra, \eg, 18, 19 (\texttt{impervious surface}), 45, 48, 52, 54 (\texttt{vegetation}) and 75 (\texttt{water}), which show a high fraction in the reconstruction. 
In the other side, some elementary spectra in this plot are infrequently used. 
It can be observed that there is generally no dependence between the height of a bar, \ie{} the number of non-zero estimated fractions for the spectra, and its overall fraction's sum for all pixels. 
A striking example is spectra 19.
Spectrum 19 in the \texttt{LibCom} library is used for more than $500$ pixel reconstructions, but its fraction's sum is lower in comparison to, e.g., spectrum 18, which is used more infrequently. 
However, \texttt{LibRed} has a more balanced composition of elementary spectra with fewer bar heights, which are close to zero.
Especially noticeable are the differences in the bar for the class \texttt{water} between \texttt{LibCom} and \texttt{LibRed}.
A water spectrum has the special characteristic that there are only small reflectance values over all bands, so that it acts almost as a linear factor. 
Because of this, it is well suited to support all reconstructions, resulting in a high overall fractions' sum. 
Its value in the \texttt{LibCom} is high with a share in over $800$ reconstructions, in comparison to \texttt{LibRed} which shows a  bar height of just over 600 and a lower fractions' sum.
We assume the reason for this is the smaller number of elementary spectra, where oftentimes a reconstruction with water spectra is helpful regardless of the presence of water in the pixel.

The results in Tab. \ref{tab:ergeb} obtained from various spectral libraries based on the labeled data set $\m X$ are more variable, indicating their dependency of a proper choice of hyperparameters such as the number of archetypes and the selection approach.
For example, the number of selected archetypes in set $\texttt{AA-M-Lin}$ is $40$, which is smaller in comparison to $\texttt{AA-M}$. 
Both sets are initialized with the mean vector of all samples in $\m X$, where the first set is extracted in the original feature space and the latter one in the kernel space.
We observed that a selection of more than $40$ archetypes in the set $\texttt{AA-M-Lin}$ means that some archetypes lie in the center of the dataset $\m{X}$, since they have the maximal distance to the previously selected archetypes.
The set $\texttt{AA-M-Lin}$ consists of $25$ impervious surface spectra, $12$ vegetation spectra, $2$ soil spectra and one water spectrum.
The total number of selected archetypes in \texttt{AA-M} is chosen to be $75$, as for the manually designed library \texttt{LibCom}. 
For all classes except the \texttt{water} class, the results obtained by \texttt{AA-M} are worse than these ones obtained by \texttt{AA-M-Lin}.
Also the results for the randomly initialized archetypal sets \texttt{AA-Rand} show only slightly better results.
Moreover, we observed that oftentimes the selected archetypal set with random initialization does not contain all land cover classes. 
Thus, as indicated by the results, a single set on selected archetypes is not suitable enough for an accurate sub-pixel quantification.
Therefore, in order to create a higher diversity of archetypes, various initializations are used for SiVM and accumulated into a larger set.
The set \texttt{AA-Full} contains $142$ different archetypes, however, the high amount of elementary spectra results in high $\operatorname{MAE}$ values.
The best result based on $\m X$ is obtained by \texttt{AA-Opt}, which is the reduced \texttt{AA-Full} set. 
This set has the best overall $\operatorname{MAE}$ and the class-wise $\operatorname{MAE}$ are in most cases comparable to the approaches based on $\m X_{\text{lib}}$.
The final number of elements in set $\texttt{AA-Opt}$ is $81.7$ on average, which is similar to the number of elements in the manually designed spectral library.  

Fig. \ref{fig:scat} underlines these findings.
For \texttt{impervious surface}, \texttt{vegetation} and \texttt{water} the results are similar for all three libraries. 
However, also for \texttt{soil} the set \texttt{AA-Opt} has large discrepancy between the true 1:1 line (dashed line) and the estimated least-squares regression line (solid line), which is more distinct than for \texttt{LibCom} and \texttt{LibRed}. 

\section{Conclusion}
The quantification of sub-pixel fractions in remote sensing images is a relevant task to assess the environmental quality, for example, in urban areas.
This paper presents a sub-pixel quantification of the urban area of Berlin, Germany, into four land cover classes \texttt{impervious surface}, \texttt{vegetation}, \texttt{soil} and \texttt{water}, using a manually designed spectral library and various kinds of automatically derived spectral libraries.
We perform the estimation of the fractions by using sparse representation with non-negativity, sparsity, and sum-to-one constraints.
We use archetypal analysis by simplex volume maximization to automatically derive the elements for a library, and apply reversible jump Markov chain Monte Carlo method to find a small, yet representative set of suitable elementary spectra.
The archetypes are extracted from HyMap data with given reference information for all land cover classes.
As our experiments suggest, the extracted archetypes are suitable to serve as spectral library for sub-pixel quantification.
Moreover, in contrast to a manually designed library, the automatically derived spectral library can be easily extracted with no or limited human user interaction, and the library is specifically adapted to the current image characteristics.
In case no reference data is given, the interpretation can be done in a fast way by assigning land cover classes to the extracted archetypes using expert knowledge.
The presented approach is flexible regarding the chosen estimation technique for the sub-pixel fractions, and can also be combined with spatial information, which may further increase the approximation ability and accuracy of the fraction estimates.

\section*{Acknowledgment}
The authors would like to thank the reviewers for their valuable comments, and Akpona Okujeni, Sebastian van der Linden and Patrick Hostert for providing the dataset. Moreover, the authors would like to thank Johannes Rosentreter and Andres Milioto for valuable discussions.

\bibliographystyle{IEEEtran}

\end{document}